\documentclass[9pt]{extarticle} 
\usepackage{spconf,amsmath,graphicx}
\usepackage{mathtools}
\usepackage{commath}
\usepackage{ctable}
\usepackage{courier} 

\title{\LARGE Self-Supervised Audio-Visual Co-Segmentation}

\name{\LARGE Andrew Rouditchenko$^{\star 1}$, Hang Zhao$^{\star 1}$,\thanks{$\star$ These authors contributed equally to this work.} Chuang Gan$^2$, Josh McDermott$^1$, Antonio Torralba$^1$}

\address{\LARGE $^1$MIT \\ \LARGE $^2$MIT-IBM Watson AI Lab \\
\texttt{\normalsize\{roudi,hangzhao,jhm,torralba\}@mit.edu, ganchuang1990@gmail.com}\\
}

\begin{document}
%
\maketitle
\vspace{10em}
\begin{abstract}
Segmenting objects in images and separating sound sources in audio are challenging tasks, in part because traditional approaches require large amounts of labeled data. In this paper we develop a neural network model for visual object segmentation and sound source separation that learns from natural videos through self-supervision. The model is an extension of recently proposed work that maps image pixels to sounds \cite{Zhao_2018_ECCV}. Here, we introduce a learning approach to disentangle concepts in the neural networks, and assign semantic categories to network feature channels to enable independent image segmentation and sound source separation after audio-visual training on videos. Our evaluations show that the disentangled model outperforms several baselines in semantic segmentation and sound source separation.

\end{abstract}

\begin{figure*}[t]
\begin{center}
\includegraphics[width=1\linewidth]{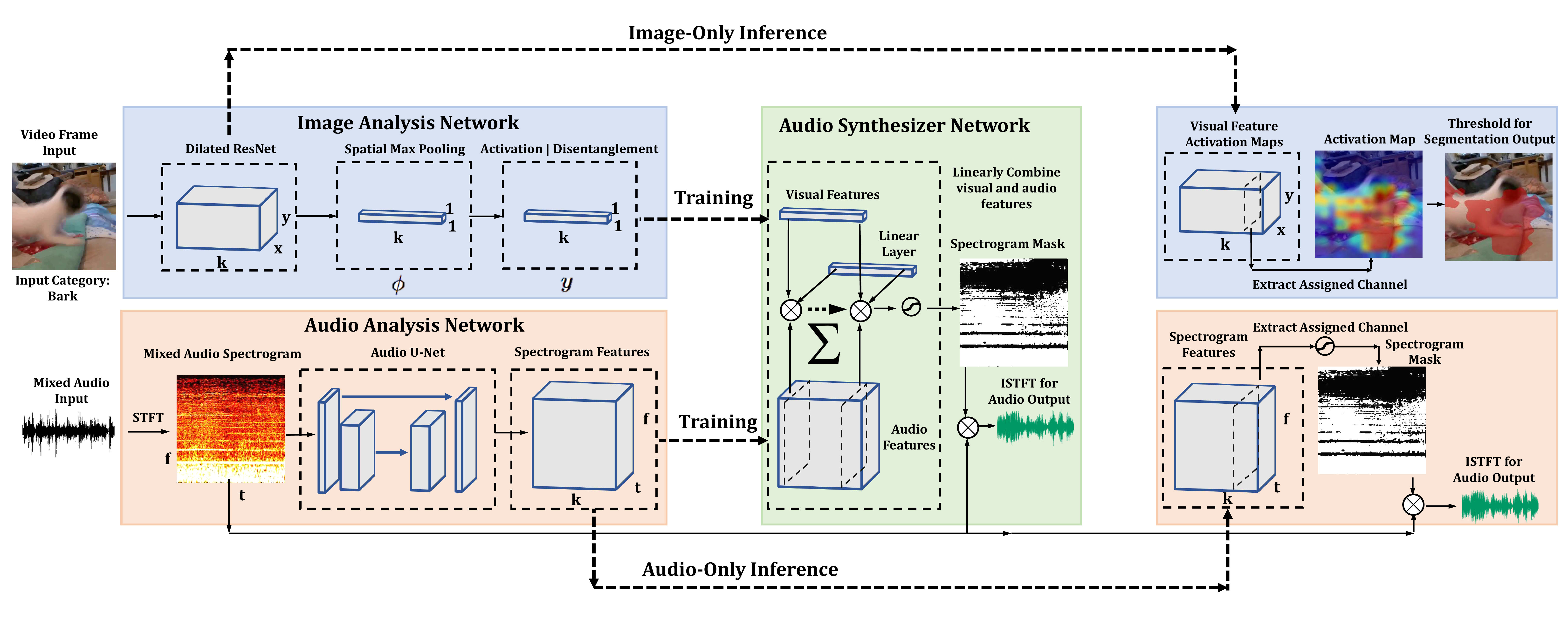}
\end{center}
\caption{Joint audio-visual training and independent image and audio inference. 
After training on synthetic mixtures of videos and assigning the dataset categories to network feature channels, the image analysis network performs image-only segmentation and the audio analysis network performs audio-only source separation. }
\label{fig:network}
\end{figure*}

\section{Introduction}
\label{sec:intro}

Semantic segmentation of images \cite{zhou2017scene,long2015fully} and sound source separation in audio \cite{bregman1994auditory, virtanen2007monaural, mcdermott2009cocktail, hershey2016deep} are two important and popular tasks in the computer vision and computational audition communities. Traditional approaches have relied on large, labeled datasets, but recent work has leveraged the natural correspondence between vision and sound to apply supervised learning without explicit labels. One approach is to use the signal or features from one modality to supervise the other. For example, \cite{aytar2016soundnet} used visual features to supervise the learning of audio networks, and \cite{owens2016ambient} used sound signals as supervision to train vision networks. Other models used sound and vision to jointly supervise each other in order to localize visual objects that make sound \cite{arandjelovic2017look,owens2018audio,pu2017audio, senocak2018learning}, and to explore the relationship between unlabelled speech and visual input \cite{kamper2019semantic,harwath2018jointly}. More recently, \cite{Zhao_2018_ECCV,ephrat2018looking,owens2018audio} used audio-visual correspondence to separate sound sources. Another key direction is to design cross-modal representations that are robust and interpretable \cite{chrupala2017representations,leidal2017learning}. Our contribution is to develop a model for audio-visual co-segmentation using videos. 

In the Mix-and-Separate framework proposed in \cite{Zhao_2018_ECCV}, neural networks are trained on videos through self-supervision to perform image segmentation and sound source separation. However, following training, the model could only be applied to videos with synchronized audio. 

Here we seek to enable a system that can perform segmentation and separation tasks using test input containing only video frames or sound mixtures. We introduce a learning approach that disentangles concepts learned by neural networks, enabling independent inference of images and audio mixtures without needing to combine visual and auditory features. The proposed learning approach relies on an activation function schedule that uses the sigmoid activation function during the training stage and a softmax activation during the fine-tuning stage, producing sparse activations that could correspond to semantic categories in the input. Following learning, semantic categories are assigned to intermediate network feature channels using labels available in the training dataset. Given a video frame or sound excerpt, the category-to-feature-channel correspondence can be used to select a particular type of source or object for resynthesis or segmentation. The disentangling thus enables both independent inference and model interpretability because the network feature channels respond sparsely to semantic concepts.   

We evaluate performance on image-only and audio-only tasks, which was not possible using the previous model. Furthermore, we substantially extend the scale of previous work \cite{Zhao_2018_ECCV} by training on a video dataset of naturally occurring audio-visual events with over 4000 videos \cite{tian2018audio}. 
The results show that we can achieve promising semantic segmentation and source source separation performance. 

\section{Approach}
\label{sec:approach}

\subsection{Self-supervised Cross-modal Training}

Our approach adopts the Mix-and-Separate framework used in~\cite{Zhao_2018_ECCV}, which first generates a synthetic sound separation training set by mixing the audio signals from two different videos, and then trains a neural network to separate the audio mixture conditioned on the visual input corresponding to one of the audio signals.  Critically, although the neural network is trained in a supervised fashion, it does not require labeled data. Thus the training pipeline can be considered as self-supervised learning.

As shown in Fig.~\ref{fig:network}, the framework we use consists of three components: an image analysis network, an audio analysis network, and an audio synthesizer network. During training, we randomly select two videos to form a synthetic training example. The image analysis network extracts visual features on a video frame from one of the videos, then uses spatial max pooling to compress the features into a visual feature vector. The audio signals of the two videos are summed and the spectrogram is extracted. An audio analysis network then processes the mixture spectrogram into audio features, where each channel contains features of different components of the input sound mixture. Finally, an audio synthesizer network combines the visual and audio features to predict a spectrogram mask to generate the audio signal for the selected video.

\subsection{Disentangling Internal Representations}
\label{sec:disentangle}

The models in \cite{Zhao_2018_ECCV} rely on synchronized video and audio as input, and thus can only perform joint audio-visual source separation, limiting their use in real applications where synchronized data are not available. Here we aim to use the image analysis network and audio analysis network independently after training, without needing the audio synthesizer network to combine the visual and audio features. 

Specifically, we design a learning schedule to disentangle the learned internal representations before the audio synthesizer network combines audio and visual features. Disentanglement is a method to create interpretable representations that enhance functionality and that make feature channels more robust to changes in other units \cite{higgins2017beta}. As shown in Fig.~\ref{fig:network}, the outputs of both the image and audio analysis networks have $K$ channels, where $K$ is larger than the number of dataset categories. Ideally, each channel would correspond to a separate concept and each input category would uniquely activate one channel. We attempt to achieve this with a learning schedule that causes the intermediate feature representations before audio-visual fusion to be sparse. 

Our technique is inspired from \cite{jang2016categorical}, who studied the effects of annealing the temperature parameter in a softmax activation function in order to push output activations towards one-hot vectors.  As the temperature parameter $T$ in the softmax activation function changes from high to low, the shape of the output distribution changes from uniform to one-hot: 
\begin{equation}
  \label{eq:softmax}
  y_k = \frac{\exp(\frac{\phi_k}{T})}{\sum^{n}_{i=1} \exp(\frac{\phi_i}{T})},
\end{equation}
where $y_k$ is the value of the $k_{th}$ visual feature channel after activation, T is the temperature, and $\phi_i$ is the value of the $i_{th}$ visual feature channel before activation. We used this idea by first training the model without imposing sparsity in the features, and then gradually changing the hyperparameters to encourage sparse and disentangled representations. The model is initially trained using a sigmoid activation on the visual feature vector $\phi$, which leads to diverse activations and helps with convergence to an initial solution. The sigmoid activation is then replaced with the softmax activation, and the temperature is gradually decreased, pushing the visual feature vector toward a one-hot vector, and causing the visual and audio feature representations to become sparse and disentangled.  Note that we did not sample from the Gumbel distribution as described in \cite{jang2016categorical}, but instead incorporated the decaying temperature parameter in the softmax activation.

\begin{table*}[tb]
\centering
\begin{tabular}{c c c c c c| c c c }
\specialrule{.2em}{.1em}{.1em}
Model Name &
\multicolumn{5}{c}{Learning Schedule} &
\multicolumn{1}{c}{Sparsity} &
\multicolumn{1}{c}{Classification} & \\

& Softmax Epochs & Initial Temp. & Decay Rate  & Decay Epochs & Final Temp. & &  \\
\hline
Baseline-Sigmoid Only & - & - & - & - & - & 0.38 &  6.30\% \\
Baseline-Softmax Only  & 25 & 1.0 & 0.3 &  10, 20 & 0.090 & 0.99 &38.7\%\\
Sigmoid \& Softmax A & 20 & 10.0 & 0.5 &  4, 8, 12, 16 & 0.625 &{0.93} &18.1\% \\
Sigmoid \& Softmax B & 20 & 1.5 & 0.75 &  4, 8, 12, 16  & 0.475 &{0.97} &37.1\%\\
Sigmoid \& Softmax C & 25 & 1.0 & 0.3 &  4, 8 & 0.090 &{0.99}  & 40.3\% \\
Sigmoid \& Softmax D & 25 & 1.0 & 0.3 &  3, 6, 9, 12 & 0.008&{0.99} &{24.0\%} \\
Sigmoid \& Softmax E & 25 & 1.0 & 0.5 &  5, 10, 15  & 0.125 &{0.99} &\textbf{45.9\%}\\
ResNet-18 Features \& SVM   & - & - & - &  - &- & {-} & 68.4\%  \\
\specialrule{.1em}{.05em}{.05em}
\end{tabular}
\vspace{-1.0em}
\caption{Classification performance and activation sparsity for the proposed model with different learning schedules and baselines. Decay Epochs indicates the epochs at which the temperature was decayed.}
\vspace{-1.0em}
\label{tab:classifcation}
\end{table*}

\begin{table}[tb]
\centering
\begin{tabular}{c | c c c c}
\specialrule{.2em}{.1em}{.1em}
Model Name & \multicolumn{2}{c}{Sound Separation} & \multicolumn{1}{c}{Seman. Seg.} \\
& SDR & SIR & IoU \\
\hline
Baseline-Sigmoid Only & 0.865 & 6.04  & 0.204 \\
Baseline-Softmax Only & 0.172 & 3.37 & 0.207 \\
Sigmoid \& Softmax A & -0.536 & 4.52 & 0.112 \\
Sigmoid \& Softmax B & 0.341 & 6.23& 0.152 \\
Sigmoid \& Softmax C & 0.716 & 6.21& \textbf{0.232} \\
Sigmoid \& Softmax D & -1.88 & 2.82& 0.205 \\
Sigmoid \& Softmax E & \textbf{1.03} & \textbf{6.37}  & 0.225\\
Nonnegative Matrix Fact. \cite{virtanen2007monaural} & 0.196 & 3.94 & - \\
Class Activation Maps \cite{zhou2016cvpr} & - & - & 0.190 \\ 
\specialrule{.1em}{.05em}{.05em}
\end{tabular}
\caption{Quantitative sound separation and semantic segmentation performance.}
\label{tab:performance}
\end{table}

\subsection{Category to Channel Assignment for Co-segmentation}
\label{sec:cosegmentation}
\vspace{-0.6em}
After training the networks without labels, we then use the category labels in the dataset to match categories to network feature channels, so that a particular type of source or object can be selected for resynthesis or segmentation. We use the validation set to compute the visual feature vector for each video and make a normalized table of these activations, which represents the cost of assigning each dataset category to each network feature channel. We then use a matching algorithm for the linear sum assignment problem \cite{kuhn1955hungarian}, which assigns each dataset category to a network feature channel. For example, the dataset category, ``cars,'' could correspond to the first network channel, ``male speech'' to the second network channel, and so forth. We can measure the validity of the assignment via classification accuracy: we measure what percentage of the input video frames in the validation set activate their assigned channel most strongly. These results are reported in Sec. \ref{sec:experiment}.

The assignment of input categories to network feature channels allows independent image and audio processing without needing the audio-synthesizer network to combine the features. In principle, one could select any activated channel and resynthesize its source signal or segmentation. Here, for evaluation purposes, we simply use the channel corresponding to the video's label in the dataset. 

For object segmentation, the last spatial max pooling layer of the image analysis network is removed to preserve activation feature maps instead of a visual feature vector.  Given an input video frame, the activation map in the channel assigned to the video's category is selected, upsampled to the input size, and thresholded to obtain a predicted segmentation. 

Given an audio mixture, the audio analysis network outputs spectrogram features in K channels. The channels assigned to the two source video categories are selected, and used as a spectrogram mask for the respective source. Each spectrogram mask is then applied to the mixture spectrogram in order to separate the corresponding sound source from the mixture.

\section{Experimental Setup}
\label{sec:experimentsetup}
\subsection{Models}

The video analysis network is a dilated variation of the ResNet-18 model \cite{he2016deep}. The dilated convolutions preserve larger visual feature activation maps, which are used after training for image segmentation. For an input video frame with size $224 \times 224$, it outputs K output activation maps of size $14 \times 14$. Spatial max pooling is then applied to compress the visual features into a visual feature vector with K channels. 

The audio analysis network is a modified U-Net \cite{ronneberger2015u} architecture. It has 7 down-convolution layers and 7 up-convolution layers with skip connections in between. For an input audio spectrogram with size $256 \times 256$, it outputs K spectrogram feature maps of size $256 \times 256$. 

The audio synthesizer network is a linear layer that is applied to combine the audio and visual features into a spectrogram mask that is multiplied element-wise with the input spectrogram. The inverse STFT is applied to the predicted magnitude spectrogram with the phase of the  input spectrogram to recover the waveform. The network outputs could be either binary or floating point masks, and we chose to use binary masks with a per pixel cross entropy loss.

\subsection{Dataset}

To train our models on a diverse set of audio-visual events, we used the Audio-Visual Event (AVE) dataset, containing 4143 videos covering 28 event categories \cite{tian2018audio}. The dataset spans categories such as cars, musical instruments, and speech, thus offering a collection that is wider in scope than other audio-visual datasets, such those limited to speech or instruments. The dataset is divided into the following splits: training (3339 videos), validation (402 videos), and test (402 videos).

The dataset was preprocessed to extract video sections containing audio-visual events,  in which the sound source is visible and the sound it produces is audible. The videos were cropped into 6-second clips, and the video frames were downsampled to 2 frames per second.  The audio signals were resampled to 11kHz and converted to spectrograms via the Short-Time Fourier Transform (STFT). The STFT used a window size of 1022 samples and a hop length of 256 samples, which resulted in $512 \times 256$ time-frequency spectrograms. For efficient model training, these spectrograms were further resampled on a on a log-frequency scale to obtain $256 \times 256$ spectrograms, which is similar to applying mel-frequency spacing. 

\begin{figure}
\begin{center}
\includegraphics[width=0.75\linewidth]{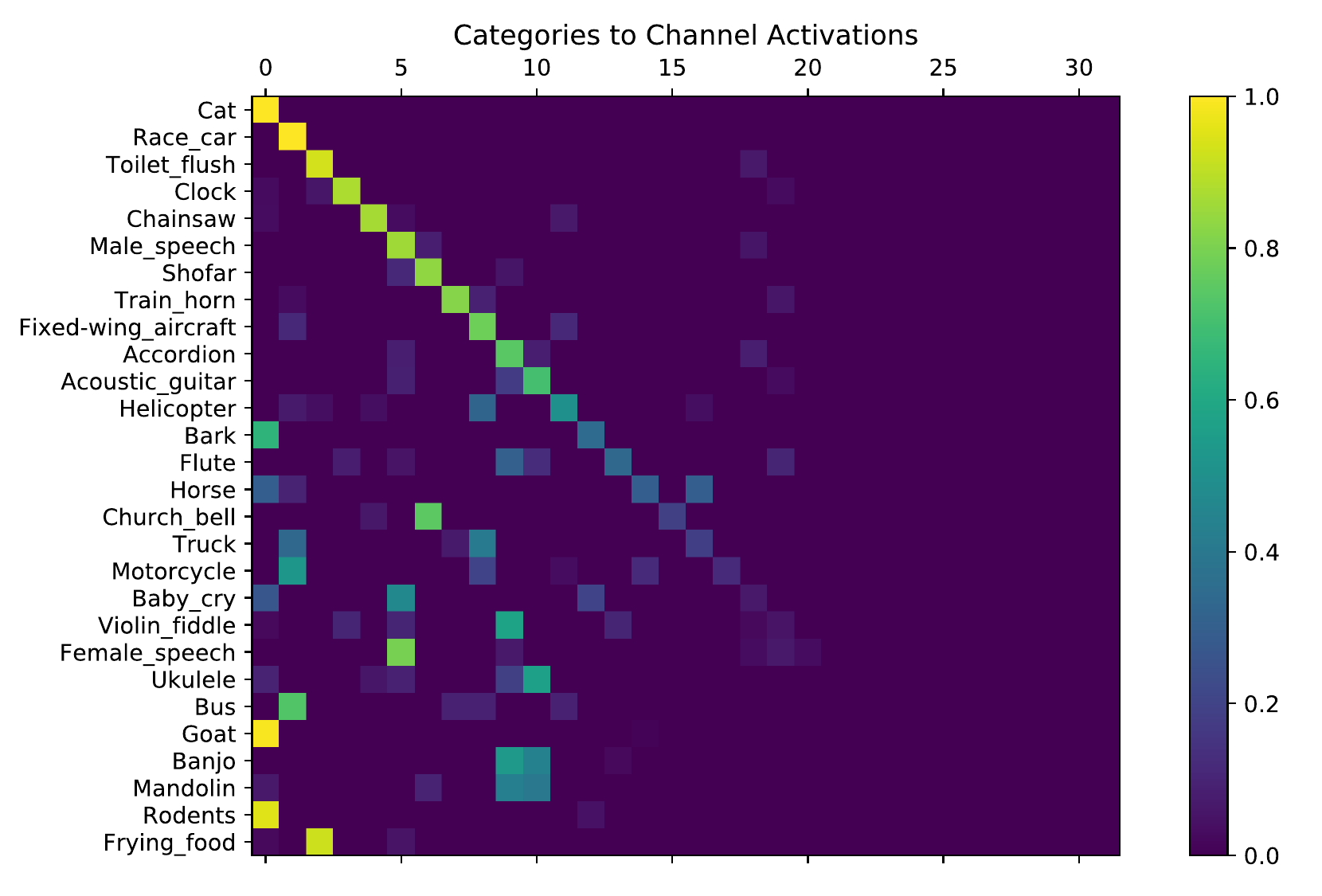}
\end{center}
\vspace{-2.0em}
\caption{Channel activations correspond to categories for the best performing model. }
\vspace{-1.0em}
\label{fig:channelactivations}
\end{figure}

\subsection{Activation Learning Schedule and Sparsity}

The learning schedule to produce sparse visual feature activations was implemented with two stages: training and fine-tuning. The training stage used a fixed sigmoid activation function and the fine-tuning stage used a softmax activation function with custom schedules for the temperature parameter. The custom schedules varied the initial temperature, the number of epochs for fine-tuning, the decay rate, and decay epochs, which proved to be important. Besides the decaying temperature, the learning rates were also reduced by a factor of five in the fine-tuning stage. We used a measure of sparsity from computational neuroscience \cite{vinje2000sparse} to evaluate the sparsity of our model activations:
\begin{equation}
    S(\mathbf{x}) = \frac{1 - (\frac{\mathbf{x} \cdot \mathbf{u}}{\norm{\mathbf{x}}~ \norm{\mathbf{u}}})^2}{1-1/K},
\end{equation}
where $\mathbf{x}$ is the channel activation, $\mathbf{u}$ is the uniform distribution, and $K$ is the number of channels in the model. We measure the sparsity of the visual feature vector after activation to quantify the extent of disentanglement. 

\section{Experimental Results}
\label{sec:experiment}
\subsection{Disentanglement and Classification}
\label{sec:classification}

In Table~\ref{tab:classifcation}, we show the performance of the proposed model (``Sigmoid \& Softmax'') with several different hyperparameter settings, as well as of the baseline models. To measure the extent of disentanglement, we evaluated the visual feature vector sparsity and classification performance on the AVE validation set. A random search over the hyperparameters was conducted to find the best performing models. We also tested different numbers of channels, $K$, and found 32 to work well and train efficiently. There are 28 categories in the AVE dataset, such that 32 channels is enough to match each category with a channel and to have extra channels for content that is not accounted for by the categories, such as silence or noise. 

The proposed model with the best hyperparameter setting, ``Sigmoid \& Softmax $E$,'' achieved a classification accuracy of $45.9\%$, significantly higher than the baseline variants of the model. To compare this result with a supervised baseline, we also trained a linear SVM on features from a ResNet-18, pre-trained on ImageNet. Although this supervised baseline achieves a higher classification accuracy of $68.9\%$, its features result from label supervision, potentially enabling fine-grained distinctions not possible using self-supervised learning. Moreover, our model has a much smaller feature vector, to enable the selection of discrete sources.   

A qualitative evaluation of the performance of the best model is shown in Fig.~\ref{fig:channelactivations}, which shows how the visual feature channels activate for different input categories. Generally speaking, each visual input category only activates one or a few channels. Some channels respond to semantically related categories, indicating that the misclassifications mostly arise due to relatively fine-grained confusions. 

The hyperparameters in the softmax fine-tuning stage proved to be important to achieve disentanglement. The softmax activation is necessary for the activations to become sparse, as shown by the low sparsity measurement from the ``Sigmoid Only'' model. The best schedule turned out to be a gradual decay from an initial temperature of 1 to about 0.01, as indicated by model ``Sigmoid \& Softmax $E$.'' The results show that ending with a temperature too high or too low can lead to suboptimal performance. 

\subsection{Source Separation}
\label{sec:separation}

\begin{figure}[t]
\begin{center}
\includegraphics[width=1.0\linewidth]{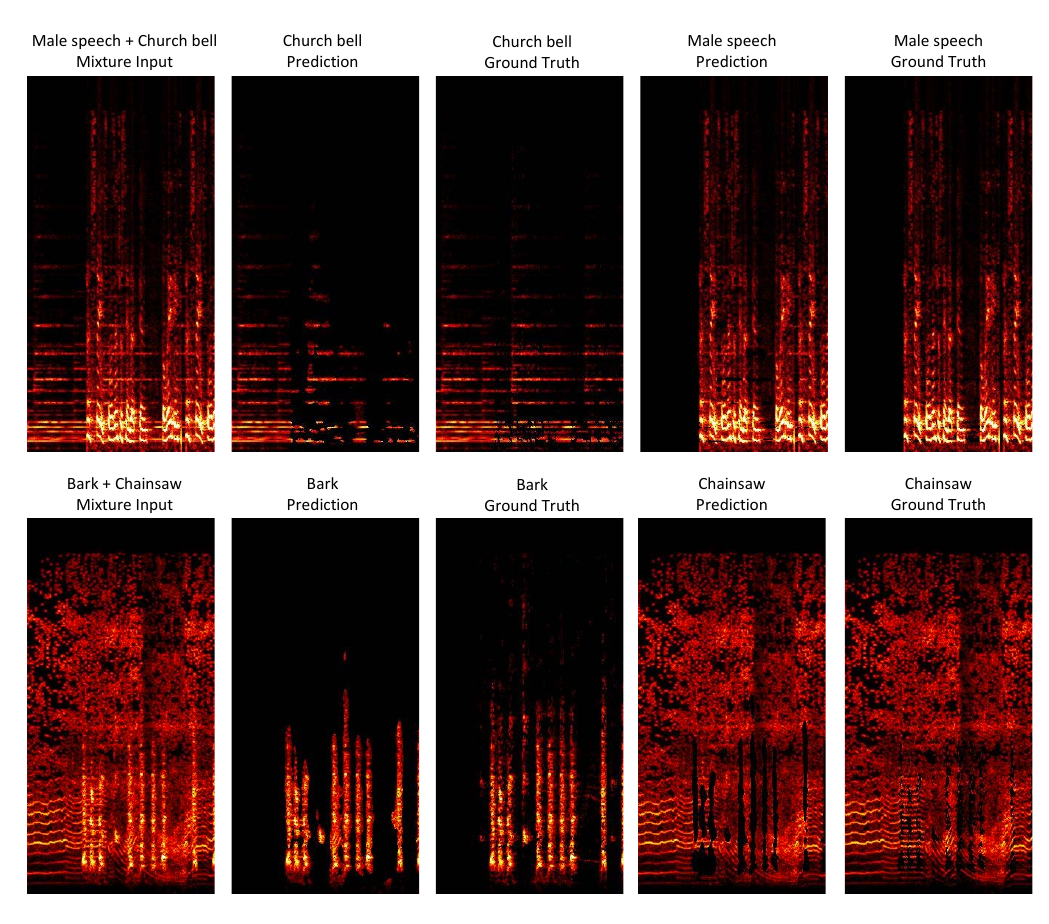}
\end{center}
\vspace{-2.5em}
\caption{Source separation results from the audio analysis network.}
\vspace{-1.0em}
\label{fig:separation}
\end{figure}

In the previous version of the model~\cite{Zhao_2018_ECCV}, source separation was only possible given synchronized audio-visual input because the network's representations of audio and video were entangled. By contrast, the current model can perform audio-only tasks following training because the sparse activations lead network feature channels to tend to correspond to semantic categories. We conducted audio-only sound source separation on the AVE test set, and show quantitative results in Table.~\ref{tab:performance} and qualitative results in Fig.~\ref{fig:separation}. The Signal to Distortion Ratio (SDR) and the Signal to Interference Ratio (SIR) are two commonly used sound source separation metrics \cite{vincent2006performance}, and were calculated using the mir-eval library \cite{raffel2014mir_eval}. We include a baseline approach of Nonnegative Matrix Factorization \cite{virtanen2007monaural}. The model which achieved the highest classification accuracy, ``Sigmoid \& Softmax $E$,'' also achieved the highest SDR and SIR. 
Qualitatively, the model succeeds in separating the sound from different sources to a large extent, which is visible in the source spectrogram recovery. 

\subsection{Semantic Segmentation}
\label{sec:segmentation}

\begin{figure}[t]
\begin{center}
\includegraphics[width=1.0\linewidth]{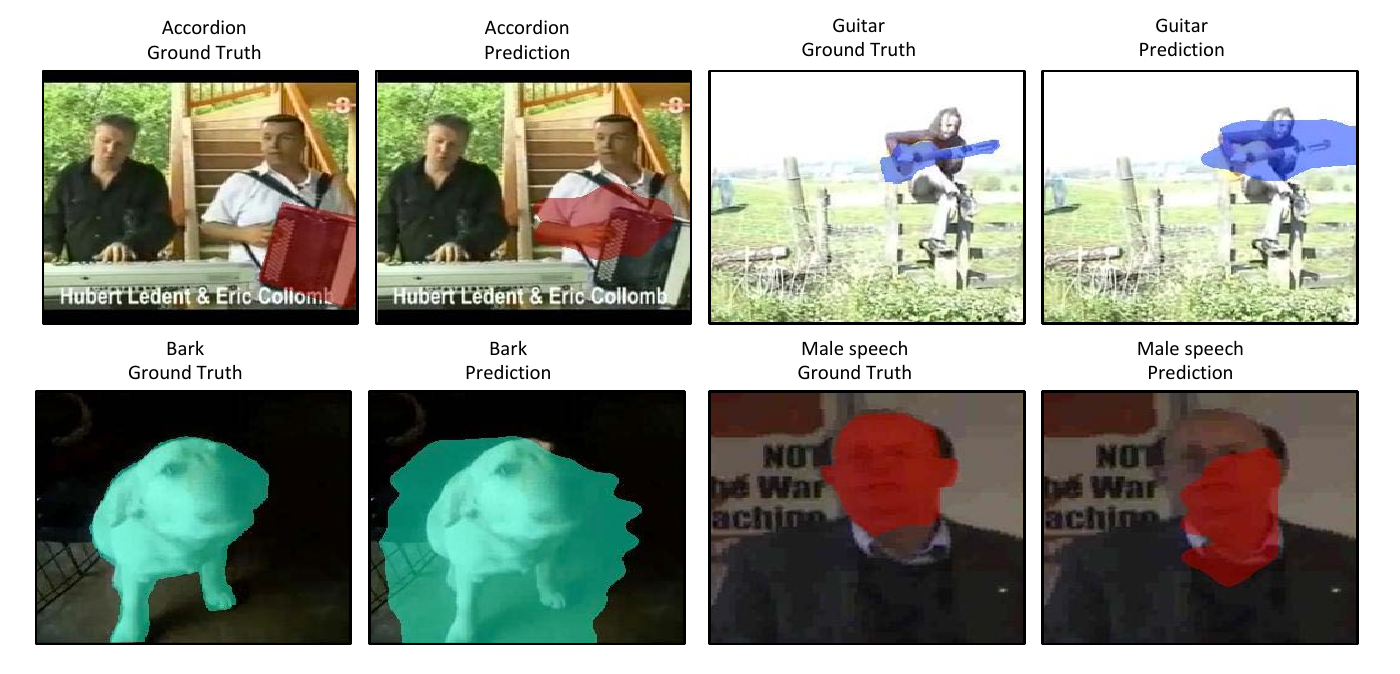}
\end{center}
\vspace{-1.5em}
\caption{Object segmentation results from the image analysis network.}
\label{fig:segmentation}
\vspace{-1em}
\end{figure}

The current model can now perform vision-only tasks following training, without the fusion of visual and audio features as in the previous version of the model \cite{Zhao_2018_ECCV}. To quantitatively evaluate the segmentation results, we labelled the middle video frame in each video from the AVE test set with polygons corresponding to the objects making sounds in the videos. The quantitative results, measured by Intersection over Union (IoU), are shown in Table~\ref{tab:performance} and qualitative results are shown Fig.~\ref{fig:segmentation}. We include a baseline approach of Class Activation Mapping \cite{zhou2016cvpr}, which is a weakly supervised method used for object localization. The best semantic segmentation performance was achieved by ``Sigmoid \& Softmax $C$,'' but the version with the highest classification performance, ``Sigmoid \& Softmax $E$,'' performed nearly as well. 
As evident in Fig.~\ref{fig:segmentation}, the boundaries of the predicted masks were often imperfect. This could result from the low resolution of the activation maps and the weak supervision used during training.

\section{Conclusion}
\label{sec:conclusion}

We developed a self-supervised audio-visual co-segmentation approach to segment visual objects and separate sound sources. The approach relied on training networks for source separation through self-supervision on a large dataset of videos. We propose a method for learning disentangled feature representations and an assignment of dataset categories to network feature channels that enables independent image segmentation and sound source separation. Experimental results on the AVE dataset show that our approach achieves promising results on semantic segmentation and source separation. 

\vfill\pagebreak

\bibliographystyle{IEEEbib}
\bibliography{refs}

\end{document}